\documentclass{article}

\usepackage{PRIMEarxiv}

\usepackage[utf8]{inputenc} 
\usepackage[T1]{fontenc}    
\usepackage{hyperref}       
\usepackage{url}            
\usepackage{booktabs}       
\usepackage{amsfonts}       
\usepackage{nicefrac}       
\usepackage{microtype}      
\usepackage{lipsum}
\usepackage{fancyhdr}       
\usepackage{xcolor}
\usepackage{graphicx}       
\graphicspath{{media/}}     

\usepackage{amsfonts,amsmath,amssymb,amsthm, mathabx}
\usepackage{mathtools}
\usepackage{thm-restate}
\usepackage{reptheorem}
\theoremfile

\usepackage{makecell}

\theoremstyle{plain}
\newtheorem{theorem}{Theorem}

\newtheorem{lemma}{Lemma}

\theoremstyle{definition}

\theoremstyle{remark}

\newcommand{\bh}{\mathbf{h}}

\newcommand{\by}{\mathbf{y}}
\newcommand{\bx}{\mathbf{x}}

\newcommand{\bA}{\mathbf{A}}
\newcommand{\bB}{\mathbf{B}}
\newcommand{\bW}{\mathbf{W}}
\newcommand{\bP}{\mathbf{P}}
\newcommand{\deri}{\mathrm{d}}

\title{LISM: Long-range Integrative State space Models via Input-Latent State Interactions}

\author{
  Cong Ma \\
  Gilbert S. Omenn Department of Computational Medicine \& Bioinformatics \\
  University of Michigan \\
  Ann Arbor, MI\\
  \texttt{congma@umich.edu} \\
  \And
  Kayvan Najarian \\
  Gilbert S. Omenn Department of Computational Medicine \& Bioinformatics \\
  University of Michigan \\
  Ann Arbor, MI\\
  \texttt{kayvan@umich.edu} \\
  \AND
  Hovhannes Baghdasaryan \\
  Institute for Informatics and Automation Problems \\
  Yerevan 0014, Armenia \\
  \texttt{hovhannes.baghdasaryan@edu.isec.am} \\
  \And
  Hrachya Astsatryan \\
  University of Michigan, Institute for Informatics and Automation Problems \\
  Yerevan 0014, Armenia \\
  \texttt{hrach@sci.am} \\
}

\begin{document}

    \maketitle

    \begin{abstract}
    State space models (SSMs) are an emerging paradigm that achieves linear-time scaling,
    however, they intrinsically suffer from ``curse of memory''. The memory of SSMs, including Mamba, decays exponentially as long as the recursive update is stable. In this work, we developed Long-range Integrative State space Models (LISM) to mitigate the curse of memory in SSMs. LISM allows the recursive update to occasionally be unstable, which enables longer memory retention, and incorporates a dynamic input-state interaction mechanism to identify inputs that should be memorized for longer periods. We theoretically establish probability bounds for the overall stability of LISM and derive a probabilistic stability condition to guide parameterization and training. We evaluated LISM's performance on multiple tasks, a synthetic Selective Copy task, two real language tasks WikiText-103 and LRA Retrieval, and an additional biological prediction task.
    LISM achieved the best performance on the Selective Copying task, WikiText-103, and the biological prediction task compared to SSM-based models and achieved comparable performance on the LRA-Retrieval task with the state-of-the-art. Beyond average performance, LISM also shows strong repeatability across runs. 
 In the two real language tasks, LISM reduced the relative performance variance (coefficient of variation) by an average of 33.3\%, achieving a standard deviation of $\pm$0.30\% on WikiText-103 compared to $\pm$0.46\% for Mamba-2. 
 The implementation of LISM is available at \url{Ananonymizedimplementation}.

    \end{abstract}
    

    \section{Introduction}
        Memorizing a long-term history is crucial for modeling sequential data in numerous domains~\cite{de200625}, such as imaging classification~\cite{shamshad2023transformers}, language prediction~\cite{devlin2019bert,conneau2019cross,raffel2020exploring}, and understanding biological sequences such as proteins and DNA~\cite{lin2023evolutionary,sanabria2024dna}. Achieving long-range memorization is a long-standing computational challenge. Classical statistical models, such as the Adaptive Integrated Moving Average (ARIMA), have been extended to identify seasonal trends that recur along a long history~\cite{tarmanini2023, hyndman2008} by incorporating a relatively simple and well-defined family of temporal patterns into the model. Recent deep learning neural models~\cite{cheng2024survey}, in particular, Recurrent Neural Networks (RNNs) \cite{medsker2001recurrent} including Long Short-Term Memory (LSTM)~\cite{hochreiter1997long} and Gated Recurrent Unit (GRU)~\cite{cho-etal-2014-properties} further enable hierarchical and complex long-term patterns by learning a hidden representation of the data that jointly models both current input and the history. Despite the success of RNNs, their accuracy requires a long runtime of model inference, in addition to manual trials and errors to deal with model instability and vanishing of gradient problems.

        Transformer models~\cite{zeng2023transformers} greatly reduce runtime compared to RNN by
        processing all tokens of a sequence in parallel while maintaining or even enhancing the ability of long-range memorization.
        Built upon a self-attention mechanism, Transformer models encode both global and local data dependencies to build a rich context representation.
        Nevertheless, efficient runtime can only be achieved during training for relatively short sequence length, as parallelization requires simultaneous access of all inputs, but Transformers have a worse runtime for long sequences or during prediction than RNN by requiring a quadratic time complexity in computing attention weights.
    
        The most recent state space models (SSMs) addressed the runtime bottleneck of both RNN and Transformers by using a linear transformation of hidden representations within each layer~\cite{gu2021s4}. A linear recursion yields a closed form representation of hidden representations across the entire sequence using convolution operations, and can be efficiently computed using Fourier transformation or parallel scan algorithms~\cite{martin2017parallelizing}. Mamba~\cite{gu2023mamba} and Mamba-2~\cite{dao2024transformers} build on these SSMs by introducing a hardware-efficient algorithm. 
        However, the latest theoretical analysis indicates that SSMs, similar to RNNs, also suffer from fast memory decay~\cite{li2022approximation,wang2023stablessm}.
    
        It remains an open question whether there exists a neural network model that achieves both efficient runtime in training and prediction phases and long-range memorization ability. In this work, we propose a novel improvement of SSM, Long-range Integrative State space Models (LISM), that achieves a superior long-range memorization with comparable runtime in both training and prediction phases. We benchmarked the accuracy of LISM using long-sequence prediction tasks including the synthetic Selective Copying task, two real language tasks, and a biological data prediction task to demonstrate its improved performance over Mamba-2 and other models.

    \section{Preliminaries}

        
        \subsection{State Space Models}
        
        SSMs are linear time-invariant systems that map input sequences \( \mathbf{x}(t) \in \mathbb{R} \) to output sequences \( \mathbf{y}(t) \in \mathbb{R} \) for $t=1, 2, \dots, L$ based on the hidden dynamics of a latent state \( \mathbf{h}(t) \in \mathbb{R}^N \). SSMs define dynamic systems as linear ordinary differential equations:
        
        \begin{equation}
        \begin{aligned}
        \mathbf{h}'(t) &= \mathbf{A} \mathbf{h}(t) + \mathbf{B} \mathbf{x}(t), \\
        \mathbf{y}(t) &= \mathbf{C} \mathbf{h}(t) + \mathbf{D} \mathbf{x}(t),
        \end{aligned}
        \label{eq:continuous_ssm}
        \end{equation}
        
        In Equation \ref{eq:continuous_ssm}, the state transition matrix  $\mathbf{A} \in \mathbb{R}^{N \times N}$ defines the autonomous system evolution, and matrix $\mathbf{B} \in \mathbb{R}^{N \times 1}$ specifies the effect of external inputs on the latent state. The output is organized by matrix $\mathbf{C} \in \mathbb{R}^{1 \times N}$ to transform the latent state, possibly aided by direct input contributions through the feedthrough matrix $\mathbf{D} \in \mathbb{R}^{1 \times 1}$.
        
        Most SSMs eliminate the direct feedthrough term by setting it to zero, leaving the output to \( \mathbf{y}(t) = \mathbf{C} \mathbf{h}(t) \). This representation skips input through middle transforms to connect directly with later layers. To convert continuous-time SSMs into a discretized model, zero-order hold \cite{astrom1984computer} methods are utilized, which assume discrete time inputs over $t_1, t_2, \dots, t_k$:
        
        \begin{equation}
        \begin{aligned}
            \mathbf{h}_k &= \overline{\mathbf{A}} \mathbf{h}_{k-1} + \overline{\mathbf{B}} \mathbf{x}_k, \\
            \mathbf{y}_k &= \mathbf{C} \mathbf{h}_k.
        \end{aligned}
        \label{eq:discrete_ssm}
        \end{equation}
        Assuming the time interval between adjacent time points is $\Delta$, the discretized matrices are transformed from the continuous matrices by
        \begin{equation}
            \overline{\bA} = \exp(\Delta \bA), \quad \overline{\bB} = (\Delta\overline{\bA})^{-1}( \exp(\Delta \bA) - I ) \Delta \overline{\bB}.
        \end{equation}
        
        Linear time-invariance in discrete SSMs enables a convenient convolutional form. The output at step \( k \) can be expressed as:
        
        \begin{equation}
            y_k = \sum_{i=0}^k \mathbf{C} \overline{\mathbf{A}}^{k-i} \overline{\mathbf{B}} x_i,
            \label{eq:ssm_convolution}
        \end{equation}
        
        which is equivalent to the convolution \( \mathbf{y} = \mathbf{K} \ast \mathbf{x}  \) with kernel \( \mathbf{K} = \{\mathbf{C}\overline{\mathbf{A}}^i\overline{\mathbf{B}}\}_{i=0}^\infty \). Instead of traditional RNNs, this formulation allows SSMs to take advantage of the hardware resources of accelerators (e.g., Graphical Processing Units or Field Programmable Gate Arrays) by computing the output sequence in parallel through one convolution operation.
        
        \subsection{Mamba}
        To enhance the capabilities of SSMs in long-range linear-time sequence modeling, Gu and Dao~\cite{gu2023mamba} leveraged several innovative techniques based on Structured State Space Models, including High-order Polynomial Projection Operator (HiPPO)-based Memory Initialization~\cite{gu2020hippo}, Selection Mechanism, and Hardware-aware Computation. By initializing the hidden state matrix $\mathbf{A}$ with a wide spectrum of eigenvalues within the stability constraint, HiPPO imposes strong induction biases that enable Mamba and other SSMs to capture temporal dependencies of various length scales. Reasoning about the mechanisms of HiPPO has also inspired other initialization strategies with good performances~\cite{orvieto2023resurrecting,agarwal2023spectral,yu2024hope}.

        Instead of using static parameters $\Delta$, $\mathbf{B}$, and $\mathbf{C}$, the selection mechanism in Mamba renders them as functions of the input by permitting $\mathbf{B}$, $\mathbf{C}$, and the time interval $\Delta$ to be input-dependent functions. The selection mechanism allows the model to dynamically regulate the flow of information into and out of the state, select pertinent inputs, and ignore irrelevant inputs. This context-sensitivity is extremely important for tasks like language modeling, where dependencies between tokens can span vastly different ranges.
        
        Hardware-aware parallel algorithm bridges the training efficiency with the inference efficiency of the recurrent mode. Instead of computing a single global convolution, the model employs a highly parallelizable scan algorithm \cite{liu2022parallel}, which is substantially optimized for modern accelerators such as GPUs or FPGAs. This allows for efficient training with parallelism and scalability, and in inference, the model can run in constant memory and time per step as an effective recurrence, similar to an RNN.
        
        The subsequent attempt by \cite{dao2024transformers} presented Mamba-2 that obtained up to eight times training speedup by leveraging 
        state-space duality. They suggested using selective SSMs that work like structured attention without sacrificing selectivity. This Mamba-2 approach enhances the utilization of tensor cores. Whereas Mamba relied exclusively on scan operations, Mamba-2 has a hybrid model that combines SSM and attention rules. This implementation enabled scaling to larger models while keeping the linear-time complexity to make SSMs faster than transformers.
        Accordingly, Mamba and Mamba-2 have demonstrated promising performance in many benchmarks and have been deployed in numerous domains~\cite{maaz2024videomamba}.
        
        \subsection{Transformer}
        Transformers~\cite{vaswani2017attention} map input sequences $\mathbf{x}_t$ to latent states $\mathbf{h}_t$ using attention mechanisms, which are further transformed by non-linear functions to output a prediction or use as input to the next transformer blocks. Transformers typically model multi-dimensional input data, but here we assume the input at each time point is a scalar to for easier comparison of the formula with SSMs. Given the linear transformation matrix $W_V \in \mathbb{R}^N$, Transformer computes the $N$-dimensional hidden state $\bh_t \in \mathbb{R}^N$ by
        \begin{equation}\label{eq:hidden_states_transformer}
          \begin{aligned}
            \bh_{t} &= \sum_{i=1}^{t} w(i, t) W_V \bx_i, \\
            \by_t &= f(\bh_t, \bx_t).
          \end{aligned}
        \end{equation}
        The attention mechanism is a linear combination of all past inputs lifted to another space with linear transformation $W_V$. The linear coefficients (attention weights) $w(i,t)$ between input at time $i$ and time $t$ is learned from their ``interaction'' after two other transformation matrices $W_Q, W_K \in \mathbb{R}^{d}$:
        \begin{equation}
            w(i, t) \propto \exp\left\{\beta (W_Q \bx_i)^\top (W_K \bx_{t}) \right\},
        \end{equation}
        where $\beta \in \mathbb{R}$ is the inverse temperature to adjust the sharpness of attention weights. After representing the following transformations using matrices $K=[W_K \bx_1, \dots, W_k \bx_t]^\top \in \mathbb{R}^{t \times d}$, $Q=[(W_Q \bx_1), \dots, (W_Q \bx_t)]^\top \in \mathbb{R}^{t \times d}$,  and $V=[ (W_V \bx_1), \dots, (W_V \bx_t) ]^\top \in \mathbb{R}^{t \times N}$, the above formula is equivalent to the typical formula of attention
        \begin{equation}
            \mathrm{Attention}(\bx_1, \dots, \bx_t) = \mathrm{Softmax}(\beta QK^\top)V.
        \end{equation}

    \section{Related Work}
        We introduce the related work in the following two aspects: theoretical study of memorization in SSM. and improvement of SSM.
        
        \subsection{Memorization in SSM}
        Some existing work has studied the memory of RNN and SSM from a theoretical perspective. For example, although RNNs are universal approximators for continuous, linear, causal, regular, and time-homogeneous functions, when fixing the model size, the approximation space of RNNs was shown to be restricted to functions with exponentially decaying memory~\cite{li2022approximation}. RNNs generally cannot easily model other functions without a great increase in model size. SSM is a specific type of RNN and its approximation space is also restricted to functions with exponentially decaying memory. Meanwhile, S4 and Mamba have improved performance than traditional RNNs, which has been shown not only to be due to the fast runtime but also to the parameterization of state transitions. Wang and Li~\cite{wang2023stablessm} studied the ``stable approximation'' property, uniform continuity of the approximation error with respect to model perturbation, and proved that parameterizing the eigenvalues of state transitions as in S4 and Mamba can achieve the stable approximation and the approximation error is robust to model perturbation. These works set the foundation of memorization capability of SSM; however, they do not suggest further improvement of memorization.

        \subsection{Improvements of SSM}
            
        Several works seek to empirically improve the memory of SSM. Block-biased Mamba~\cite{yu2026block} improved memory focusing on the formula of input-dependent time interval $\Delta$ and showed that allowing $\Delta$ from different input dimensions to follow different input-dependent functions helps improve memory. Although this improvement leads to a slower decay rate, the model still has a theoretically exponential decay of memory. ReMamba~\cite{yuan2024remamba} proposes a workaround by a two-phase training strategy: identify potential important input elements in the first phase and highlight these inputs to run Mamba a second time, a similar idea to Retrieval-Augmented Generation (RAG)~\cite{lewis2020retrieval}. Despite the soundness of the idea, the actual performance depends on the accuracy of identified important inputs, and the complexity in training may prevent application to different types of data when training from scratch is needed. GatedDeltaNet~\cite{yang2025gated} modifies the state-transition matrix and introduces gating~\cite{yang2024gatedlinearattention} to erase memory and the delta rule~\cite{schlag2021linear,yang2024parallelizing} to incorporate input-specific information into the state-transition matrix beyond $\Delta$. However, the formula in the delta rule does not effectively enhance memory, as we show in the next section.
    
        Transformer, by avoiding recursions in deriving latent states, still provides an out-of-the-box framework to circumvent the ``curse of memory''. Some works combine SSM and Transformers and create hybrid models to have a longer memory than SSM alone~\cite{Xiao2024-cb,Xu2024-li,Zuo2024-yz}. However, the disadvantages of long runtimes of Transformers are also introduced. Griffin~\cite{de2024griffin} and the subsequent Recurrentgemma~\cite{botev2024recurrentgemma} seek to reduce the runtime of hybrid models by using linear recurrences and local attention; however, local attention cannot retain long-range memory by ignoring distant inputs.

    \section{Improving the long-range Memory of SSM by LISM}

        \subsection{Memory of SSM and Transformers}
            Previous work~\cite{li2022approximation,wang2023stablessm} uses a continuous-time memory function to study the behavior of RNN and Mamba. Here, we introduce a simplified version of memory in the discrete-time formula and generalize the memory analysis to Transformers. Given input $\bx_t$ at time $t$, we define the memorization capability of the latent state $\bh_{t+k}$ ($k > 0$) of the input as:
            \begin{equation}
                \mathcal{M}(t+k, t) = \frac{\deri \bh_{t+k}}{\deri \bx_t}.
            \end{equation}
            A zero norm of $\mathcal{M}(t+k, t)$ indicates the latent state at $t+k$ is not affected by the input at time $t$, or the memory is lost. A large norm of $\mathcal{M}(t+k, t)$ indicates $\bh_{t+k}$ is largely affected by a small change in $\bx_t$, or $\bx_t$ is memorized at time $t+k$. Under this discrete memory definition, we re-state the exponentially decay memory property of SSM, including Mamba, which determines a variable time gap $\Delta_t$ between consecutive inputs $\bx_t$ and $\bx_{t+1}$ that is learned from the inputs.
            \begin{theorem}
                Given an input $x_t$, the $L_2$ norm of memory decays exponentially with the rate $\lambda_1$ per unit time, where $\lambda_1$ is the magnitude of the largest eigenvalue of $\bA$ and is non-positive for stable SSM.
                \begin{equation}
                        \| \mathcal{M}_{SSM}(t+k+1, t) \|_2 \leq
                          \exp(\lambda_1 \Delta_{t+k+1}) \| \mathcal{M}_{SSM}(t+k, t) \|_2.
                \end{equation}
            \end{theorem}
            The exponentially decaying memory of SSM results from the intrinsic formulation of the model: the stability of SSM over infinite inputs requires the real parts of eigenvalues of transition matrix $\bA$ to be upper bounded by 0 (or the magnitudes of eigenvalues of discretized $\bar{\bA}$ upper bounded by 1), however, a slowly decaying memory requires the eigenvalues of $\bA$ to be close to or even above 0.

            The memory of a single attention layer of transformers between latent states $\bh_{t+k}$ at time $t+k$ and the input $x_t$ at time $t$ is
                 \begin{equation}\label{eq:memory_transformer}
                        \mathcal{M}_{\text{transformer}}(t+k, t) = \sum_{i=1}^{t+k-1}  \frac{\deri w(i, t+k)}{\deri \bx_t} W_V \bx_i + 
                        w(t, t+k) W_V.
                \end{equation}
            See Appendix for the detailed derivation. Despite being a complicated form, $\mathcal{M}_{\text{transformer}}(t+k, t)$ does not decay exponentially with the time gap $k$ if the interaction $w(t,t+k)$ is large between time $t$ and time $t+k$. Transformers circumvent the limit memory problem by avoiding recursion of latent states and determining whether a past input $\bx_t$ should be memorized by an ``interaction'' between the past input and current information computed through the attention mechanism. However, despite the flexibility induced by the pairwise ``interaction'', the time complexity for computing all pairwise ``interactions'' is much higher than computing recursions, especially when the computation cannot be parallelized during prediction.

        \subsection{Long-range Integrative State space Models (LISM)}
            We propose a mitigation to the curse of memory in SSM by combining the ideas of SSM and Transformers. The motivation for our proposed method is based on the following argument: let $\lambda_1$ be the magnitude of the largest eigenvalue of the discretized state transition matrix $\bar{\bA}$ in SSM, when $\lambda_1 < 1$, SSM is stable but memory exponentially decays; when $\lambda_1 \geq 1$, SSM is unstable but allows longer memory. Past research has not considered relaxing the bound of $\lambda_1$. Nevertheless, if $\lambda_1$ is allowed to be variable, $\lambda_1 < 1$ most of the time but occasionally exceeds 1, the SSM can still be stable, but certain inputs will be memorized for a longer time. We incorporate the ``interaction'' between past input and current latent state to determine which inputs need to be memorized for longer. We propose the following recursive update of latent states model, which we call Long-range Integrative State space Models (LISM):
            \begin{equation}\label{eq:hidden_state_translation_raw}
                \bh_t = \bar{\bA} \bh_{t-1} + \bar{\bB}_t \bx_t + (\bh_{t-1}^\top \bW \bx_t)(\tilde{\bB} \bx_t).
            \end{equation}
            The first two terms on the right-hand side are the same as in SSM, where we allow $\bar{\bB}_t$ to be determined by the input $\bx_t$ as in Mamba model. $\bar{\bA}$ is the discretization of the state transition matrix $\bA$ with time interval $\Delta$, and the above formula can also be adapted to the case where the time interval $\Delta_t$ is dynamically determined by the input $\bx_t$. For simplicity of notation, we use $\bar{\bA}$ to represent both fixed time interval and input-dependent time interval cases. The third term encodes the ``interaction'' between the previous information $\bh_{t-1}$ and the current input $\bx_t$. The strength of the interaction is $\bh_{t-1}^\top \bW \bx_t$, the similarity between $\bh_{t-1}$ and $\bx_t$ in a projected space by $\bW$. The hidden state will be shifted along the direction $\tilde{\bB} \bx_t$ weighted by the interaction strength, where $\tilde{\bB}$ resembles the ``value'' matrix in transformers. Mamba model is a special case of our new formulation by fixing $\tilde{\bB}$ to a zero matrix. After reorganizing the terms, we obtained the following equivalent formula.
            \begin{equation}\label{eq:hidden_state_translation}
                \bh_t = (\bar{\bA} + \tilde{\bB} \bx_t \bx_t^\top \bW^\top) \bh_{t-1} + \bar{\bB}_t \bx_t.
            \end{equation}
            The unrolled form of the hidden state update yields the following.
                \begin{equation}\label{eq:unrollform_translation}
                        \bh_{t} ={}
                        \left(\prod_{i=1}^t
                        (\bar{\bA} + \tilde{\bB} \bx_i \bx_i^\top \bW^\top)\right) \bh_0
                         + \sum_{i=1}^t \left( \prod_{j=i+1}^{t}
                        (\bar{\bA} + \tilde{\bB} \bx_j \bx_j^\top \bW^\top) \right) \bar{\bB}_i \bx_i .
                \end{equation}
            where the $\prod_i^j$ notation for matrix multiplication denotes the left matrix multiplication starting with the index at the bottom ($i$) and ending with the index at the top $j$. The matrix multiplication in the unrolled form allows efficient computational using the parallel scan algorithm as in S5~\cite{smith2023simplified} and Mamba. 
            
            Focusing on the Single-Input Single-Output (SISO) system used by Mamba, where inputs are assumed to be scalars $\bx_t \in \mathbb{R}$, we derive the memory of LISM as follows.
            \begin{makethm}{lemma}{lem:lism_memory}
                The memory of LISM~\eqref{eq:hidden_state_translation_raw} is 
                \begin{equation}
                    \begin{aligned}
                        & \mathcal{M}(t+k, t) = \mathbf{S}_{t+1,t+k} (2x_t \tilde{\bB} \bW^\top) \mathbf{S}_{1,t-1}  \bh_0 \\
                        & \qquad + \sum_{i=1}^{t-1}  \mathbf{S}_{t+1,t+k} (2x_t \tilde{\bB} \bW^\top) \mathbf{S}_{i+1,t-1}   \bar{\bB}_i x_i \\
                        & \qquad + \mathbf{S}_{t+1,t+k}  \bar{\bB}_t,
                    \end{aligned}
                \end{equation}
                where $\mathbf{S}_{a, b} = \prod_{i=a}^{b} (\bar{\bA} + \tilde{\bB}_i x_i^2 \bW^\top)$.
            \end{makethm}
            See Appendix for the detailed derivation. We further derive the following decay rule of LISM, that is, the decay of $\mathcal{M}(t+k+1,t)$ compared to $\mathcal{M}(t+k,t)$:
            \begin{equation}
                \mathcal{M}(t+k+1,t) = (\bar{\bA} + \tilde{\bB}_{t+k+1} x_{t+k+1}^2 \bW^\top) \mathcal{M}(t+k,t).
            \end{equation}
            The magnitude of eigenvalues of $(\bar{\bA} + \tilde{\bB}_{t+k+1} x_{t+k+1}^2 \bW^\top)$ may exceed 1 for certain inputs, allowing the model to memorize these inputs for longer time and breaking the exponential decay constraint. We empirically computed the memory $\mathcal{M}(t+k,t)$ with respect to a fixed $t$ using random Gaussian inputs and observed that it increases at certain inputs, in contrast to the exponential decay (Figure~\ref{fig:memory_comparison}). 

            GatedDeltaNet~\cite{yang2025gated} has a similar hidden state update formula, however, it constrains the eigenvalues to below 1 and thus still suffers from the exponentially decaying memory. Specifically, it uses the hidden state update formula of $\bh_t = \alpha_t \bh_{t-1}(I - \beta_t f(\bx_t) f(\bx_t)^\top) + \beta_t \mathbf{v}_t f(\bx_t)^\top$, where $\alpha_t, \beta_t \in (0,1)$ and $f(\bx_t)$ is usually a linear transformation of input $\bx_t$. The memory function of GatedDeltaNet has a similar form as Lemma~\ref{lem:lism_memory}, however, the state transition matrix is $\alpha_t(I - \beta_t f(\bx_t) f(\bx_t)^\top)$ and the eigenvalues cannot exceed $\alpha_t \leq 1$, thus failing to improve the long-range memorization capability.
        
            \begin{figure}[!htbp]
              \centering
              \includegraphics[width=0.8\textwidth]{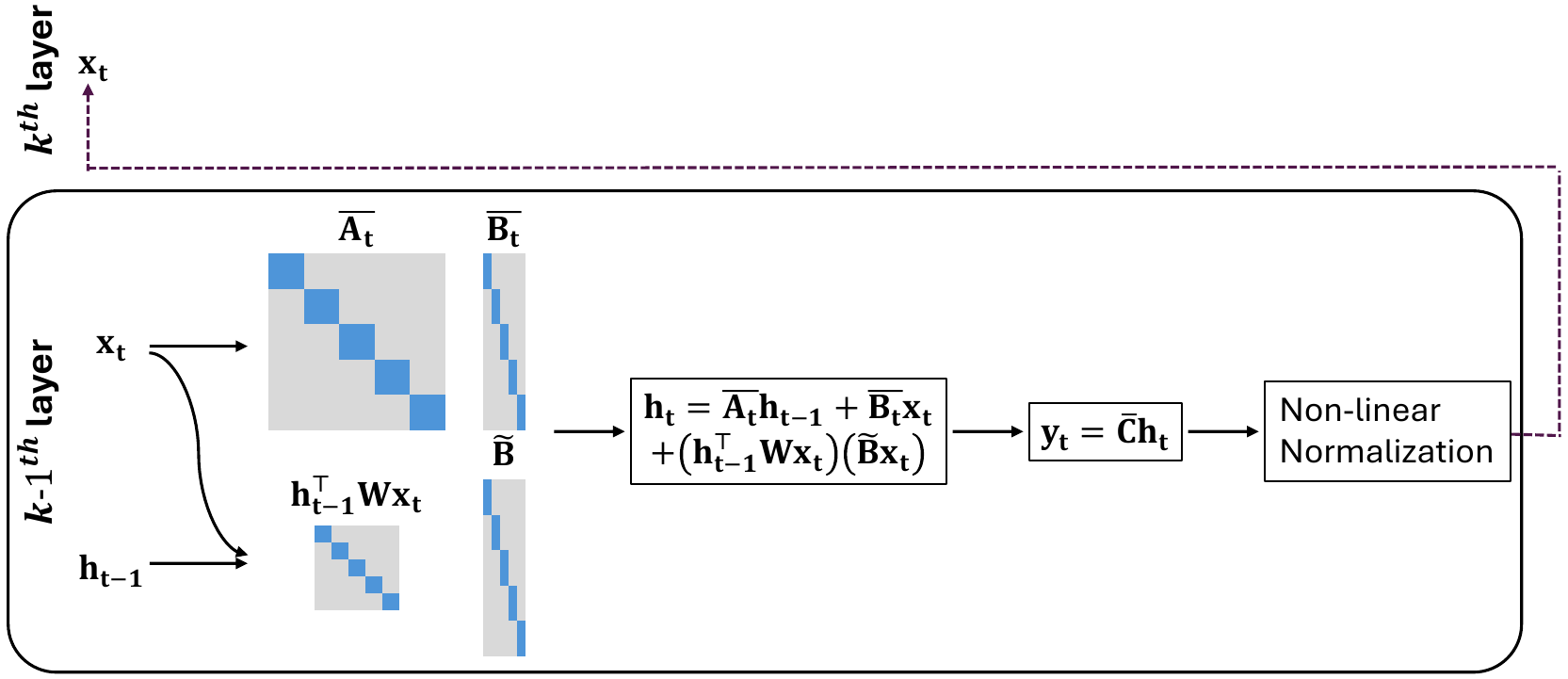}
              \caption{Architecture of LISM showing dynamic input-state interaction mechanism}
              \label{fig:lism-architecture}
            \end{figure}

        \subsection{Stability of LISM}
            The eigenvalues of the input-dependent state transition $\bar{\bA} + \tilde{\bB} \bx_t \bx_t^\top \bW^\top$ of LISM is allowed to occasionally exceed 1 for some inputs $\bx_t$, the stability threshold. However, the system is still overall stable if this case does not occur frequently. In this section, we assume the inputs $\bx_t$ are identically distributed following a Gaussian distribution and analyze the probability that the system is overall stable. Note that the Gaussian distribution is a reasonable assumption for many types of sequential data with proper preprocessing. Although the derivation assumes an independent assumption, the stability bound is effective in practice with real language and biological data as we demonstrate in experiments. 
        
            To further simplify the model, we additionally assume $\tilde{\bB} x_t^2 \bW^\top$ can be diagonalized simultaneously with $\bar{\bA}$ besides following SISO system. The co-diagonalization assumption indicates that $\tilde{\bB}$ and $\bW$ are scalings of the same eigenvectors of $\bar{\bA}$. Let the eigenvalues and their corresponding eigenvectors of $\bar{\bA}$ be $\lambda_1, \dots, \lambda_H$ and $\bP = [\bP_1, \dots, \bP_H]$, accordingly, $\bar{A} = \bP \Lambda \bP^\top$ where $\Lambda$ is the diagonal matrix of the eigenvalues. Note that we do not assume $\lambda_1, \dots, \lambda_H$ to be sorted. Without loss of generality, we assume $\tilde{\bB}$ and $\bW$ are scalings of the eigenvector $\bP_H$ in the decomposition, and $\tilde{\bB}\bW^\top = \gamma \bP_H \bP_H^\top$. Under this setup, the hidden state update of LISM is
            \begin{equation}\label{eq:practical_lism_update}
                \bh_t = (\bP\Lambda \bP^\top + \gamma x_t^2 \bP_H \bP_H^\top) \bh_{t-1} + \bar{\bB}_t x_t.
            \end{equation}
            We focus on the case where the time interval $\Delta_t=1$ is fixed across inputs, but the stability can be generalized to the time-dependent time interval case. The transition matrix can then be written as
            \begin{equation}\label{eq:state_transition_eigen}
                \bar{\bA} + \tilde{\bB} x_t^2 \bW^\top = \bP \begin{pmatrix} \lambda_1 & & &\\ & \lambda_2 & & \\ & & \ddots & \\ & & & \lambda_H + \gamma x_{t}^2 \end{pmatrix} \bP^\top
            \end{equation}
            The first $H-1$ eigenvalues, $\lambda_1,\dots, \lambda_{H-1}$, must be no greater than 1 for the system to be stable. We proved the following theorems to upper bound stability of state transitions related to the last eigenvalue. 
            %
            \begin{makethm}{lemma}{lem:stabilityexpression}
                Assuming $0 < |\lambda_1|, \dots, |\lambda_{H-1}| < 1$, the hidden state of LISM at time $t$, $\bh_t$, is finite if $\sum_{i=1}^t \prod_{j=i+1}^t |\lambda_H + \gamma x_{j}^2|$ is finite.
            \end{makethm}
            
            \begin{makethm}{theorem}{thm:probabilistic_bound_statibility}
                Assuming the inputs $x_1, x_2, \dots$ are identically distributed under a standard Gaussian distribution, and $\lambda_H$ is the complex eigenvalue of $\bar{\bA}$, then there exists a pair of positive values $(\nu, b)$ such that
                \begin{equation}
                      P\left(\prod_{i=1}^t |\lambda_H + \gamma x_{i}^2| \geq e^z ( |\lambda_H| + \gamma)^t \right)\leq
                      \left\{
                      \begin{aligned}
                      & e^{-\frac{z^2}{2 t\nu^2}} \quad \text{if } 0 \leq z \leq \frac{t\nu^2}{b} \\
                      & e^{-\frac{t}{2b}} \quad \text{if } z > \frac{t\nu^2}{b}.
                      \end{aligned} \right.
                \end{equation}
            \end{makethm}
        
            See Appendix for the proof. With fixed $| \lambda_H | + \gamma < 1$ and increasing values of $t$, we further have
            \begin{equation}\label{eq:final_bound}
                    P\left(\sum_{i=1}^t\prod_{j=i+1}^t |\lambda_H + \gamma x_{j}^2| \leq e^z \frac{1 - (|\lambda_H| + \gamma)^t}{1 - |\lambda_H| - \gamma} \right)  \geq
                    1 - C(b),
            \end{equation}
            for a small constant $C(b)$ related to $b$. (See Appendix for detailed derivation.) Therefore, it is unlikely that the hidden states become unstable as long as $| \lambda_H | + \gamma < 1$.

    \section{Experiments}
        We evaluated the performance of LISM on the synthetic selective copy task and two real language tasks, WikiText-103 and Retrieval benchmark of Long Range Arena (LRA)~\cite{tay2021long}. We benchmarked against the following architectures that represent the dominant paradigms in sequence modeling: Mamba~\cite{gu2023mamba}, Mamba-2~\cite{dao2024transformers}, Transformer~\cite{vaswani2017attention}.
        We did not find the github implementation of Block-biased Mamba~\cite{yu2026block} and therefore could not include it in our benchmarking. However, LISM model targets a different aspect from Block-biased Mamba and could in theory be combined for further improvements. The implementation of GatedDeltaNet~\cite{yang2025gated} is incompatible with our GPUs, thus we did not compare directly with GatedDeltaNet. Instead, we implemented a special version of LISM to mimic the update of GatedDeltaNet by constraining $\gamma$ parameters to non-positive values, which we refer to LISM-nonpos. We evaluated the test accuracy, wall time per training step, wall time per inference time, and GPU memory of each method.

        \subsection{Implementation of LISM}
            We implemented the hidden state transition of \eqref{eq:practical_lism_update} based on the implementation of Mamba. Since large (resp. small) eigenvalues are designed for long-range (resp. short-range) memorization, we added the $\gamma x^2$ term to the largest eigenvalue of $\bar{\bA}$ to better preserve long-range memory. We adapted the implementation of Mamba for LISM to take advantage of advanced techniques applied to other parameters, including HiPPO initialization of eigenvalues and the selective scan mechanism to allow input-dependent parameters. To guarantee the stability condition of the parameter $\gamma$, $\lambda_H + \gamma < 1$, we do not directly parameter $\gamma$ but instead parameter an unscaled version $\gamma^o$ and compute $\gamma = \mathrm{tanh}(\gamma^o) (1 - \lambda_H)$. The eigenvalue $\lambda_H$ is input-dependent according to $\bar{\bA}_t = \exp(\Delta_t A)$ with input-dependent time interval $\Delta_t$. HiPPO initializes a broad range of eigenvalues and provides a strong inductive bias; consequently, the relative order of eigenvalues is often preserved throughout training. We computed all hidden states according to \eqref{eq:state_transition_eigen} using the parallel scan algorithm~\cite{blelloch1990prefix,gu2023mamba}, for which we implemented custom CUDA kernels based on those of Mamba.

        \subsection{Selective copy task}
            The selective copy task requires a model to read an input sequence containing both tokens to be memorized and noise tokens, and then to output a sequence containing only the tokens to be memorized in the same order as input. The synthetic task was used by S4~\cite{gu2021s4}. We used the same input sequence length (4096) and vocabulary size (16) as in S4, but enlarged the number of tokens (three settings of 64, 128, and 256) to memorize in order to better evaluate the long-range memorization capability. We randomly generated 1K test samples and 300K training samples based on github~\cite{zou2024selective} while guaranteeing that the training set and the test set do not have any overlapping samples.

            We used two layers of LISM model (\verb|L=2|) with an internal dimension of 512 (\verb|d=512|), which has 3.41 million parameters. We used the same architecture for Mamba and Mamba-2 and adjusted transformer-based models to have approximately the same model size. Since LISM uses an additional parameter $\gamma$ and has more parameters than Mamba-2 under the same architecture, we additionally tested LISM with a smaller number of internal dimensions (497) to match the model size of Mamba and Mamba-2, which we refer to as LISM match (or LISM-M). The details of model architectures are in Table~\ref{tab:configs}. When embedding the input tokens, we included a positional embedding and concatenated with token embedding for SSM-based models such that the concatenated embedding size matched the assumed embedding dimension (512). SSM and RNN models typically do not need positional embedding for real language data because the semantics in the language implicitly carry some positional information, and the hidden state recursion implicitly encode positional information to some degree. However, the synthetic task randomly generates tokens from the vocabulary that do not have semantic meanings, and thus comparing transformers that naturally require positional encoding with SSM without position encoding yields an unfair result for SSM-based models. Therefore, we explicitly included positional encoding for SSM-based models for a fair comparison. 
            
            We trained SSM-based models on V100 GPUs under limit a maximum number of 37.5K training steps and a maximum 690 minutes of wall time. Since Mamba-2 runs faster per training step, we increased the maximum number of training steps of Mamba-2 proportionally so that it uses approximately the same walltime of training compared to other methods. All models used a batch size of 32.
            The learn rate is set to $10^{-4}$ for SSM-based models and $10^{-3}$ for transformer-based models with a warmup phase of 1000 training steps. The randomly simulated data has an unsmooth likelihood landscape, therefore we clipped the gradient that exceeds 50 gradient norm, in contrast to the standard gradient norm when training on real language data.

            \begin{table}[htbp]
                \centering
                \footnotesize
                \setlength{\tabcolsep}{5pt}
                \renewcommand{\arraystretch}{1.2}
                \begin{tabular}{|p{2.1cm}|p{3cm}|c|}
                \hline
                \textbf{Model} & \textbf{Architecture Config} & \textbf{Params} \\
                \hline
                \textbf{LISM} & $d{=}512$, $L{=}2$ & $\sim$3.41M  \\
                \hline
                \textbf{LISM-M} & $d{=}497$, $L{=}2$ & $\sim$3.22M  \\
                \hline
                \textbf{LISM-nonpos} & $d{=}512$, $L{=}2$ & $\sim$3.41M  \\
                \hline
                \textbf{Mamba} & $h{=}512$, $L{=}2$ & $\sim$3.41M  \\
                \hline
                \textbf{Mamba-2} & $d{=}512$, $L{=}2$ & $\sim$3.22M  \\
                \hline
                \textbf{Transformer} & $d{=}512$, $L{=}2$, heads: 8 & $\sim$3.20M  \\
                \hline
                \end{tabular}
                \caption{Standardized model configurations}
                \label{tab:configs}
            \end{table}

\begin{figure}[!ht]
                \centering
                \includegraphics[width=0.5\linewidth]{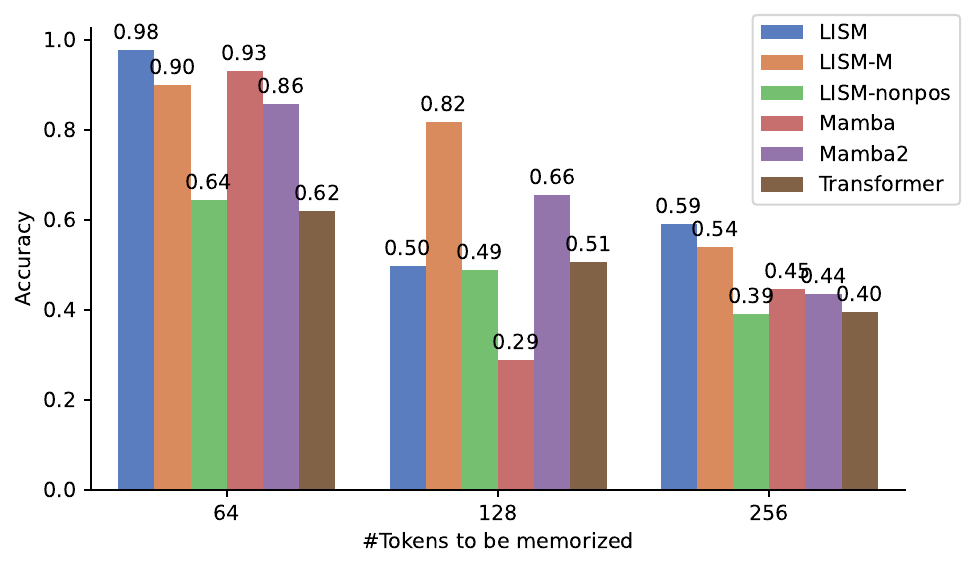}
                \caption{Accuracy of LISM and other models in Selective Copy task.}
                \label{fig:accuracy_selective_copy}
            \end{figure}

            LISM does not have numerical overflow problems in this task, demonstrating the probabilistic bound in Theorem~\ref{thm:probabilistic_bound_statibility} effectively guarantees the stability of the model in practice. LISM achieved the highest test accuracy on two of the three settings (Fig.~\ref{fig:accuracy_selective_copy}). In the hardest task (256 tokens to memorize), LISM exceeded the second best model by 14\% accuracy. The performance of LISM is not merely due to an increased number of parameters than other SSM-based models, because a parameter-matched setup in LISM-M achieved higher accuracies than other methods on the harder settings of 128 and 256 tokens and a comparable accuracy on the 64-token setting. 

            The accuracy of LISM relies on allowing the $\gamma$ parameter to span both positive and negative ranges, for which a positive $\gamma$ theoretically breaks the exponential decay constraint. If constraining the $\gamma$ parameter to be only non-positive as assumed by GatedDeltaNet, the accuracy greatly decreases on all settings, which is shown by the LISM-nonpos bar in Fig.~\ref{fig:accuracy_selective_copy}.

            Benefited from the SSM framework, LISM has an efficient runtime and GPU memory (Table~\ref{tab:computing_efficiency}). We benchmarked the walltime required for training and inferring each batch and the GPU memory for each method using the A40 GPU. LISM requires 1.45 times of the runtime and 1.28 times of the GPU memory of Mamba-2 in order to compute the additional term in the state transition matrix. But the runtime and GPU memory are still much smaller than Transformer.

            \begin{table}[htbp]
                \centering
                \footnotesize
                \setlength{\tabcolsep}{5pt}
                \renewcommand{\arraystretch}{1.2}
                \begin{tabular}{|p{1.6cm}|p{1.7cm}|p{1.7cm}|p{1.4cm}|}
                \hline
                \textbf{Model} & \textbf{training walltime (sec)} & \textbf{inference walltime (sec)} & \textbf{GPU memory (MB)} \\
                \hline
                \textbf{LISM} & 1.0908 & 0.2836 & 22872.5 \\
                \hline
                \textbf{LISM-M} & 1.0734 & 0.3006 & 22211.15 \\
                \hline
                \textbf{Mamba} & 0.7834 & 0.1765  & 17823.72 \\
                \hline
                \textbf{Mamba-2} & 0.7529 & 0.1554 & 17816.79 \\
                \hline
                \textbf{Transformer} & OOM & OOM & OOM \\
                \hline
                \end{tabular}
                \caption{Computing efficiency benchmarked on A40 GPU with batch size of 32.}
                \label{tab:computing_efficiency}
            \end{table}

        \subsection{Real language tasks}

            \textbf{WikiText-103.} We evaluated LISM and LISMv2 on the WikiText-103 language modeling benchmark~\cite{merity2016pointer}. WikiText-103 contains a collection of wikipedia articles with 103 million unique words. It has been used as a standard real language benchmarking task for predicting next tokens. We used GPT-2 tokenizer to encode text sequences into token sequences with a vocabulary size of 50.26K and predicted the next token for each sequence of length $l_{\max} = 4096$. We used 16 layers with an internal dimension of $\mathrm{d_{model}} = 1024$ for SSM-based models and trained on four A40 GPUs with an effective batch size of 120 and a fixed training time limit of 4.5 hours. The fixed training time limit aligns with our previous task evaluation and allows us to better evaluate the effectiveness of training in a resource-limited setting. 
            
            LISM achieved the highest accuracy (lowest perplexity) in all SSM-based models and out-performed Mamba and Mamba-2 (Table~\ref{tab:wikitext103}), and meanwhile achieved the most robust accuracy in the three random initializations demonstrated by the small standard deviation of perplexity. Transformer is known to have the best performance in this task~\cite{gu2021s4}, indicating that long-range dependency is crucial for accurate prediction and direct pairwise interactions of inputs serve as the most effective way of long-range memorization. LISM closed the gap to Transformer by enhancing long-range memorization and enabling the eigenvalues of the hidden state transition to exceed 1 for a small subset of inputs. We observed that when constraining $\gamma \leq 0$ in LISM to mimic GatedDeltaNet, LISM-nonpos also achieved better performance than Mamba and Mamba-2, suggesting that selectively forgetting certain inputs faster may also be effective in this task. However, the high performance of the selective forgetting strategy in LISM-nonpos was not consistent across random initialization and did not robustly provide a low perplexity. In contrast, LISM generates consistently better performance across random seeds, resulting in a standard deviation of 0.059 and coefficient of variance of  $\pm$0.30\%, which is much lower than LISM-nonpos and lower than the next best performing method Mamba-2 (standard deviation 0.092 and coefficient of variation $\pm$0.46\%.

            \begin{table}[htbp]
                \centering
                \footnotesize
                \setlength{\tabcolsep}{3pt}
                \renewcommand{\arraystretch}{1.2}
                \begin{tabular}{|c|c|c|c|c|}
                \hline
                \textbf{Model} &  \textbf{Perplexity Avg (Std)} & \textbf{Model size} \\
                \hline
                \textbf{Transformer}  & 17.827 (0.211) & 217.41M \\
                \hline\Xhline{2\arrayrulewidth}
                \textbf{LISM} & \textbf{19.91} (0.059) & 209.63M \\
                \hline
                \textbf{LISM-nonpos} & \underline{19.94} (0.150) & 209.63M \\
                \hline
                \textbf{Mamba}  & 20.72 (0.130) & 209.60M \\
                \hline
                \textbf{Mamba-2}  & 20.11 (0.092) & 204.86M \\
                \hline
                \end{tabular}
                \caption{WikiText-103 test perplexity. Bold text highlights the best accuracy across small-size models, and underline highlights the second best.}
                \label{tab:wikitext103}
            \end{table}

            \textbf{LRA Retrieval.} As LRA repository was archived, we could not obtain data for most tasks in LRA for benchmarking, but only obtained Retrieval task data from an unofficial huggingface site~\cite{retrieval_lra}. Retrieval task requires the model to read two articles from ACL Anthology Network and predict whether the two articles share a common citation. Following the standard processing, we tokenized each article into a length-4000 sequence with a vocabulary size of 98 and passed both articles to the same model architecture. The final binary label was predicted by combining the model output of both articles and passing that through a three-layer MLP classifier. We set the number of layers as 6 for all SSM models and set the internal dimension as 128 for Mamba, Mamba-2, and LISM, and 256 for S4. We set the number of layers as 4 and the internal dimension as 128 following the training setup in \cite{amos2024never}. We trained each model using batch size 64 and applied an 8-hour time limit for training and evaluated the accuracy of binary prediction on the test set. 
            
            LISM achieved the second highest accuracy on this task, demonstrating its robustness to various language prediction tasks (Table~\ref{tab:retrieval_task}). Within a limit of 8 hours, LISM reached a test accuracy of 90.03\% across three random initializations, which is higher than Mamba, S4, Transformer. Mamba-2 achieved a slightly higher accuracy than LISM but with a larger standard variation in its accuracy across random initializations. LRA prediction relies on a combined hidden state from both articles, and thus does not require a substantial length of memorization. Consequently, the long-range memorization of LISM does not yield a substantial benefit to this task, while Mamba-2 benefits from its faster training and larger number of training steps within the time limit. Amos et al.~\cite{amos2024never} reported that pre-training of Transformers to predict the next tokens helped to learn a better parameter configuration, which leads to a more accuracy binary prediction accuracy of Retrieval task through subsequent training. We evaluated this training strategy under the same time limit by evenly splitting training time in pre-training and final training and observed that the test accuracy (Transformer+SPT) is much higher than without pretraining but is still lower than LISM and SSM-based models. Note that the accuracies are not directly comparable with previous reported LRA Retrieval task~\cite{amos2024never}: here we evaluate the training efficiency under a fixed time budget for all models, while previous benchmarking fixed the number of epochs instead of the wall time and allowed the training wall time to be drastically different across models. We chose the 8-hour time limit because the fastest models (Mamba and Mamba-2) reach the recommended number of training steps (50K) within the time limit, representing the minimum computation cost for this task. This does not contradict the reported higher accuracy of Transformers in other work, as Transformers were typically trained with over 400K steps and much longer wall time than this setting. With a limited time budget, training SSM-based models is more effective than training Transformers, and LISM achieved the top two training effectiveness on this task.

            \begin{table}[htbp]
                \centering
                \footnotesize
                \setlength{\tabcolsep}{5pt}
                \renewcommand{\arraystretch}{1.2}
                \begin{tabular}{|p{2.4cm}|p{2.5cm}|p{1.1cm}|p{1.5cm}|}
                \hline
                \textbf{Model} & \makecell{\textbf{Accuracy}\\mean (std)} & \textbf{Size} & \textbf{Forward FLOPs/1e9} \\
                \hline
                \textbf{LISM} & 90.03\% (0.15\%) & 725.4K & 5.16 \\
                \hline
                \textbf{LISM-M} & \underline{90.31\%} (0.23\%) & 682.9K & 4.83 \\
                \hline
                \textbf{Mamba} & 88.31\% (2.69\%) & 723.8K  & 5.14 \\
                \hline
                \textbf{Mamba-2} & \textbf{90.71\%} (0.22\%) & 688.8K & 5.14 \\
                \hline
                \textbf{S4} & 89.00\% (0.38\%) & 1.11M & 16.949 \\
                \hline
                \textbf{Transformer} & 77.31\% (0.91\%) & 793K & 35.945 \\
                \hline
                \textbf{Transformer+SPT} & 84.97\% (1.40\%) & 793K & 35.945 \\
                \hline
                \end{tabular}
                \caption{Accuracy on LRA Retrieval task. Bold text highlights the best accuracy across small-size models, and underline highlights the second best.}
                \label{tab:retrieval_task}
            \end{table}    

        \subsection{Biological task}
            We applied LISM to a small size of the biological task of AlphaGenome~\cite{avsec2026advancing} to demonstrate its generalizability to other real tasks other than language modeling. AlphaGenome seeks to simultaneously predict 5930 human features and 1128 mouse features from the DNA sequence that consists of ``A'', ``C'', ``T'', ``G'' in the vocabulary. AlphaGenome uses a multi-task training framework that uses a common base architecture based on U-net and transformer with multiple prediction heads for each prediction feature, eventually containing 450 million trainable parameters in total. We focused on 10 human features for prediction and evaluated the accuracy of LISM and other models with a much smaller parameter size.

            The 10 human features are the coverages of RNA-seq, ATAC-seq, ChIP-seq H3K4me1, ChIP-seq H3K27me3, ChIP-seq H3K9ac across two tissue types colonic mucosa and lung. These features include multiple types of crucial molecular features; tissues are selected to avoid missing modality. We used the same training loss (a weighted sum of Poisson and Multinomial negative log-likelihoods) and the same test accuracy evaluation metric (Pearson correlation) as in AlphaGenome and previous work~\cite{linder2025predicting}. AlphaGenome is trained on each 1Mb; however, to reduce computation resources, we train all models on 10Kb windows. Chromosome 19 is used as the test set and the other chromosomes are used to train the model. We used a 6-layer model for LISM, Mamba, and Mamba2 with dimension \verb|d=512| and a 6-layer model with dimension 192 and 8 attention heads for Transformer. We similarly set an 8 hour time limit for training all models.

            LISM showed the highest Pearson correlation on average across the 10 features in the test set (Table~\ref{tab:dna_task}) and is the closest to the state-of-the-art accuracy from AlphaGenome. Meanwhile, LISM has a much smaller model size than AlphaGenome and was trained with a much shorter time with a single A40 GPU compared to AlphaGenome which was trained on 512 TPU cores and 64 H100 GPUs for three days. The results show the average across three random seeds for all methods, except for Mamba because its training experienced bad gradient steps and resulted in near 0 Pearson correlation for two random seeds. In contrast, the training of LISM is more stable although it allowed the eigenvalues of hidden state transition matrix to occasionally exceed 1. Transformer achieved the lowest accuracy, potentially related to the phenomenon that transformer cannot be robustly trained when training from scratch~\cite{amos2024never}, especially when the training time has a tight budget.

            \begin{table*}[htbp]
                \centering
                \footnotesize
                \setlength{\tabcolsep}{5pt}
                \renewcommand{\arraystretch}{1.2}
                \begin{tabular}{|p{1.8cm}|p{0.8cm}|p{0.85cm}|p{0.85cm}|p{0.8cm}|p{1.1cm}|p{1.25cm}|p{1.0cm}|p{0.8cm}|p{0.8cm}|p{1.1cm}|p{1.25cm}|p{1.0cm}|}
                \hline
                \textbf{Model} & \textbf{Model size} & \textbf{Avg} (std) & \textbf{Colonic-RNA} & \textbf{Colonic-ATAC} & \textbf{Colonic-H3K4me1} & \textbf{Colonic-H3K27me3} & \textbf{Colonic-H3K9ac} & \textbf{Lung-RNA} & \textbf{Lung-ATAC} & \textbf{Lung-H3K4me1} & \textbf{Lung-H3K27me3} & \textbf{Lung-H3K9ac}\\
                \hline
                \textbf{AlphaGenome} & 450M & 0.341 (NA) & 0.271 & 0.238 & 0.427 & 0.290 & 0.457 & 0.420 & 0.173 & 0.476 & 0.252 & 0.404 \\
                \hline\Xhline{2\arrayrulewidth}
                \textbf{LISM}        & 2.53M& \textbf{0.274} (0.0013) & \textbf{0.089} & \underline{0.432} & \underline{0.333} & \textbf{0.237} & \textbf{0.347} & \underline{0.107} & \textbf{0.323} & \textbf{0.398} & \textbf{0.179} & \textbf{0.289 }\\
                \hline
                \textbf{Mamba}       & 2.52M& 0.258 (NA) & 0.083 & 0.397 & 0.324 & 0.232 & 0.322 & 0.086 & 0.299 & 0.389 & 0.171 & 0.281 \\
                \hline
                \textbf{Mamba-2}      & 2.41M& \underline{0.270} (0.0035) & \textbf{0.089} & \textbf{0.433} & \textbf{0.334} & \textbf{0.237} & \underline{0.335} & \textbf{0.113} & \underline{0.320} & \underline{0.393} & \underline{0.173} & \underline{0.275} \\
                \hline
                \textbf{Transformer} & 2.66M& 0.232 (0.0037) & 0.079 & 0.386 & 0.281 & 0.214 & 0.291 & 0.093 & 0.279 & 0.353 & 0.130 & 0.219 \\
                \hline
                \end{tabular}
                \caption{Pearson correlation between predicted and observed coverage on the test set of the biological task. Bold text highlights the best accuracy across small-size models, and underline highlights the second best.}
                \label{tab:dna_task}
            \end{table*}

    \section{Conclusion}
        In this paper, we proposed LISM, a state-space model that is capable of long-range memorization. LISM introduces a novel interaction term between hidden states and input, allowing the model to selectively memorize the input for longer according to the relevance of the input. Contrary to Mamba and other existing sequential SSM models, where eigenvalues of state transitions are constrained to be below 1, LISM allows the eigenvalues to exceed 1 for a small number of inputs, breaking the exponential decaying constraint underlying previous SSMs.
        Benchmarking LISM with previous SSMs and transformers of similar model size and wall training time, we observed LISM obtained highest accuracy in multiple tasks, including the synthetic selective copy task, the real language task WikiText-103, and a biological DNA to epigenetic prediction task.

        Beyond average performance, LISM also shows strong repeatability across runs, with consistently smaller standard deviations than most competing methods. This is especially important in small-data settings common in medical and biomedical applications, where model reliability is critical. 
        Compared with Transformer-based models, LISM's structured state-transition parameterization, constrained by the stability condition in Theorem~\ref{thm:probabilistic_bound_statibility}, appears less sensitive to initialization: perplexity std on WikiText-103 ($\pm$0.059) is $\sim$3.6$\times$ lower than Transformer's ($\pm$0.211), and accuracy std on LRA Retrieval ($\pm$0.15\%) is $\sim$6$\times$ lower ($\pm$0.91\%).

        Looking ahead, LISM can be further improved in several ways. LISM currently still consumes more time for each training step compared to Mamba and Mamba-2, and its computing efficiency can be further improved by implementing a fused cuda kernel as in Mamba. LISM currently assumes that the projection vectors of the interaction term align with the eigenvectors of $\bar{\bA}$, while enabling a flexible projection vector will further enhance the model expressiveness.

        Overall, LISM shows that the dynamic input-state interaction in the state-space model provides a strong paradigm for long-range memorization of SSM. The idea is motivated by the pairwise input interaction of Transformer, and thus LISM represents a new research direction on integrating the advantages of SSM and Transformer into a coherent framework with runtime efficiency.
    
    \bibliographystyle{unsrt}
    \bibliography{reference}

\newpage
\onecolumn

\setcounter{figure}{0}
\renewcommand{\thefigure}{A\arabic{figure}}

\appendix
\begin{figure}[!htbp]
    \centering
    \includegraphics[width=0.5\textwidth]{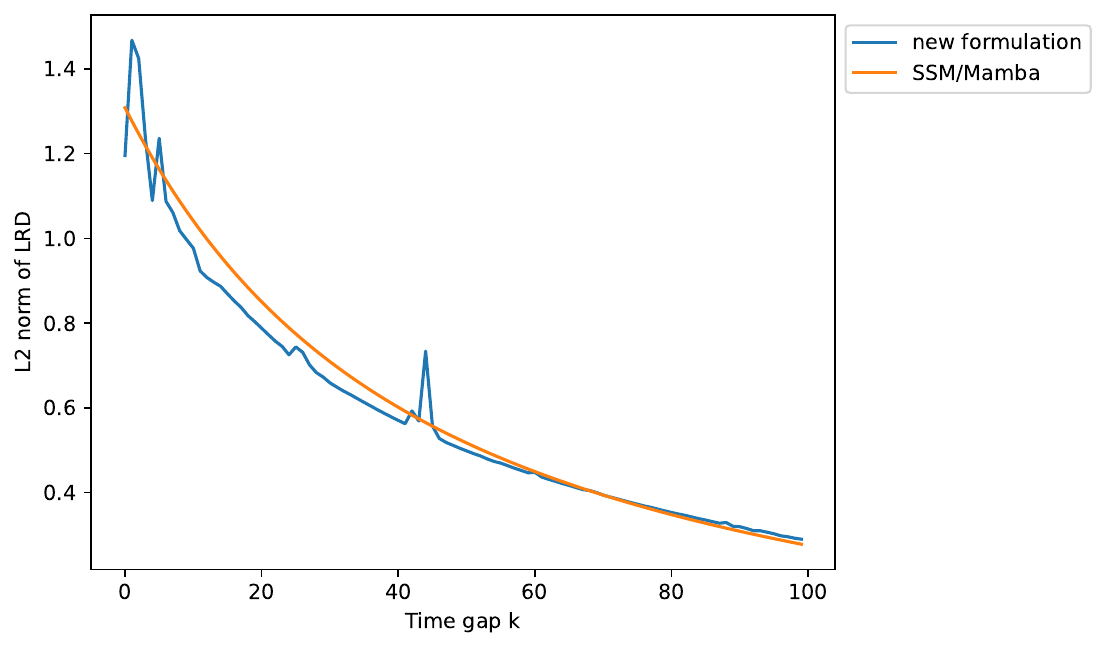}
    \caption{Memory $\mathcal{M}(t+k, t)$ of new formula and SSM using simulated matrices and inputs with $k=1, \dots, 100$.}\label{fig:memory_comparison}
\end{figure}

\section{Derivation of memory of Transformer~\eqref{eq:memory_transformer}}
    Using the hidden state formula of Transformer~\eqref{eq:hidden_states_transformer}, we derive the memory as
    \begin{equation*}
        \begin{aligned}
            \mathcal{M}_{\text{transformer}}(t+k, t) &= \frac{\deri \bh_{t+k}}{\deri \bx_t} \\
            &= \frac{\deri \big( \sum_{i=1}^{t+k} w(i, t+k) W_V \bx_i \big) }{\deri \bx_t} \\
            &= \sum_{i=1}^{t+k} \frac{\deri w(i, t+k)}{\deri \bx_t} W_V \bx_i + w(i, t+k) W_V
        \end{aligned}
    \end{equation*}

\section{Derivation of the unrolled form of LISM hidden states~\eqref{eq:unrollform_translation}}
    We derive the form by induction. Treating the initial hidden state $\bh_0 = \mathbf{0}$ as a zero vector, the theorem claims the next hidden state is
    \begin{equation*}
      \begin{aligned}
        \bh_1 &= \left(\prod_{i=1}^1 (\bar{\bA} + \tilde{\bB} \bx_i \bx_i^\top \bW^\top)\right) \bh_0 + \sum_{i=1}^1 \left( \prod_{j=i+1}^{1} (\bar{\bA} + \tilde{\bB} \bx_j \bx_j^\top \bW^\top) \right) \bar{\bB}_i \bx_i \\
        &= (\bar{\bA} + \tilde{\bB} \bx_1 \bx_1^\top \bW^\top) \bh_0 + \left( \prod_{j=2}^{1} (\bar{\bA} + \tilde{\bB} \bx_j \bx_j^\top \bW^\top) \right) \bar{\bB}_1 \bx_1 \\
        &= (\bar{\bA} + \tilde{\bB} \bx_1 \bx_1^\top \bW^\top) \bh_0 + \bar{\bB}_1 \bx_1.
      \end{aligned}
    \end{equation*}
    This form aligns with the hidden state transition in \eqref{eq:hidden_state_translation}.

    Assuming the $(t-1)$-th hidden state satisfy the form, 
    \begin{equation*}
        \bh_{t-1} = \left(\prod_{i=1}^{t-1} (\bar{\bA} + \tilde{\bB} \bx_i \bx_i^\top \bW^\top)\right) \bh_0 + \sum_{i=1}^{t-1} \left( \prod_{j=i+1}^{t-1} (\bar{\bA} + \tilde{\bB} \bx_j \bx_j^\top \bW^\top) \right) \bar{\bB}_i \bx_i.
    \end{equation*}
    The $t$-th hidden state according to the state transition \eqref{eq:hidden_state_translation} is
    \begin{equation*}
      \begin{aligned}
        \bh_t &= (\bar{\bA} + \tilde{\bB} \bx_t \bx_t^\top \bW^\top) \bh_{t-1} + \bar{\bB}_t \bx_t \\
        &= (\bar{\bA} + \tilde{\bB} \bx_t \bx_t^\top \bW^\top) \left\{ \left(\prod_{i=1}^{t-1} (\bar{\bA} + \tilde{\bB} \bx_i \bx_i^\top \bW^\top)\right) \bh_0 + \sum_{i=1}^{t-1} \left( \prod_{j=i+1}^{t-1} (\bar{\bA} + \tilde{\bB} \bx_j \bx_j^\top \bW^\top) \right) \bar{\bB}_i \bx_i \right\} + \bar{\bB}_t \bx_t \\
        &= (\bar{\bA}_t + \tilde{\bB} \bx_t \bx_t^\top \bW^\top) \left(\prod_{i=1}^{t-1} (\bar{\bA} + \tilde{\bB} \bx_i \bx_i^\top \bW^\top)\right) \bh_0 + \sum_{i=1}^{t-1} (\bar{\bA} + \tilde{\bB} \bx_t \bx_t^\top \bW^\top) \left( \prod_{j=i+1}^{t-1} (\bar{\bA} + \tilde{\bB} \bx_j \bx_j^\top \bW^\top) \right) \bar{\bB}_i \bx_i + \bar{\bB}_t \bx_t
      \end{aligned}
    \end{equation*}
    Because we use the $\prod$ to denote the left matrix multiplication, we can further simplify the above formula:
    \begin{equation*}
      \begin{aligned}
        \bh_t &= \left(\prod_{i=1}^{t} (\bar{\bA} + \tilde{\bB} \bx_i \bx_i^\top \bW^\top)\right) \bh_0 + \sum_{i=1}^{t-1} \left( \prod_{j=i+1}^{t} (\bar{\bA} + \tilde{\bB} \bx_j \bx_j^\top \bW^\top) \right) \bar{\bB}_i \bx_i + \bar{\bB}_t \bx_t \\
        &= \left(\prod_{i=1}^{t} (\bar{\bA} + \tilde{\bB} \bx_i \bx_i^\top \bW^\top)\right) \bh_0 + \sum_{i=1}^{t-1} \left( \prod_{j=i+1}^{t} (\bar{\bA} + \tilde{\bB} \bx_j \bx_j^\top \bW^\top) \right) \bar{\bB}_i \bx_i + \left( \prod_{j=t+1}^{t} (\bar{\bA} + \tilde{\bB} \bx_j \bx_j^\top \bW^\top) \right) \bar{\bB}_t \bx_t \\
        &= \left(\prod_{i=1}^{t} (\bar{\bA} + \tilde{\bB} \bx_i \bx_i^\top \bW^\top)\right) \bh_0 + \sum_{i=1}^{t} \left( \prod_{j=i+1}^{t} (\bar{\bA} + \tilde{\bB} \bx_j \bx_j^\top \bW^\top) \right) \bar{\bB}_i \bx_i.
      \end{aligned}
    \end{equation*}
    Hence, when $\bh_{t-1}$ satisfy the formula in the theorem, $\bh_t$ obtained using the recursive hidden state transition will also satisfy the formula in the theorem. By this proof of induction, the unrolled hidden state formula in the theorem holds for recursive transition~\eqref{eq:hidden_state_translation}.

\section{Proof of Lemma~\ref{lem:lism_memory}}
    \repthm{lem:lism_memory}
    \begin{proof}
        Applying the hidden state formula~\eqref{eq:unrollform_translation} in memory $\mathcal{M}(t+k, t) = \frac{\deri \bh_{t+k}}{\deri \bx_t}$, we obtain
        \begin{equation*}
            \begin{aligned}
                \mathcal{M}(t+k, t) &= \frac{\deri}{\deri x_t} \left\{  \left(\prod_{i=1}^{t+k}
                        (\bar{\bA} + \tilde{\bB} x_i^2 \bW^\top)\right) \bh_0 + \sum_{i=1}^{t+k} \left( \prod_{j=i+1}^{t+k}
                        (\bar{\bA} + \tilde{\bB} x_j^2 \bW^\top) \right) \bar{\bB}_i x_i  \right\} \\
                        &= \left(\prod_{i=t+1}^{t+k}
                        (\bar{\bA} + \tilde{\bB} x_i^2 \bW^\top)\right) \frac{\deri}{\deri x_t} (\bar{\bA} + \tilde{\bB} x_t^2 \bW^\top) \left(\prod_{i=1}^{t-1}
                        (\bar{\bA} + \tilde{\bB} x_i^2 \bW^\top)\right) \bh_0 \\
                        & + \sum_{i=1}^{t+k} \left( \frac{\deri}{\deri x_t} \prod_{j=i+1}^{t+k}
                        (\bar{\bA} + \tilde{\bB} x_j^2 \bW^\top) \right) \bar{\bB}_i x_i + \sum_{i=1}^{t+k} \left( \prod_{j=i+1}^{t+k}
                        (\bar{\bA} + \tilde{\bB} x_j^2 \bW^\top) \right) \frac{\deri}{\deri x_t} (\bar{\bB}_i x_i)
            \end{aligned}
        \end{equation*}
        The last term $\sum_{i=1}^{t+k} \left( \prod_{j=i+1}^{t+k} (\bar{\bA} + \tilde{\bB} x_j^2 \bW^\top) \right) \frac{\deri}{\deri x_t} (\bar{\bB}_i x_i)$ has only one nonzero term since $\frac{\deri}{\deri x_t} (\bar{\bB}_i x_i) = 0$ whenever $i \neq t$. Accordingly, the last term can be expressed as $\left( \prod_{j=t+1}^{t+k} (\bar{\bA} + \tilde{\bB} x_j^2 \bW^\top) \right) \bar{\bB}_t$.
        
        Denoting $\mathbf{S}_{a, b} = \prod_{i=a}^{b} (\bar{\bA} + \tilde{\bB} x_i^2 \bW^\top)$, $\mathbf{S}_{a, b}$ is only relevant to $\bx_t$ when $a \leq t \leq b$. Under this notation, we simplify the above expression as
        \begin{equation*}
            \begin{aligned}
                \mathcal{M}(t+k, t) &= \mathbf{S}_{t+1, t+k} \frac{\deri}{\deri x_t} (\bar{\bA} + \tilde{\bB} x_t^2 \bW^\top) \mathbf{S}_{1, t-1} \bh_0 + \sum_{i=1}^{t+k} \left( \frac{\deri}{\deri x_t} \prod_{j=i+1}^{t+k}
                        (\bar{\bA} + \tilde{\bB} x_j^2 \bW^\top) \right) \bar{\bB}_i x_i + \mathbf{S}_{t+1, t+k} \bar{\bB}_t \\
                        &= \mathbf{S}_{t+1, t+k} (2x_t \tilde{\bB} \bW^\top) \mathbf{S}_{1, t-1} \bh_0 + \sum_{i=1}^{t+k} \left( \frac{\deri}{\deri x_t} \prod_{j=i+1}^{t+k}
                        (\bar{\bA} + \tilde{\bB} x_j^2 \bW^\top) \right) \bar{\bB}_i x_i + \mathbf{S}_{t+1, t+k} \bar{\bB}_t
            \end{aligned}
        \end{equation*}
        In the middle term, $\bar{\bA} + \tilde{\bB} x_j^2 \bW^\top$ does not involve $x_t$ when $i \geq t$, thus $\sum_{i=t}^{t+k} \left( \frac{\deri}{\deri x_t} \prod_{j=i+1}^{t+k}(\bar{\bA} + \tilde{\bB} x_j^2 \bW^\top) \right) \bar{\bB}_i x_i = 0$. For $i < t$, $\frac{\deri}{\deri x_t} \prod_{j=i+1}^{t+k}(\bar{\bA} + \tilde{\bB} x_j^2 \bW^\top) = \prod_{j=t+1}^{t+k}(\bar{\bA} + \tilde{\bB} x_j^2 \bW^\top) \frac{\deri}{\deri x_t} (\bar{\bA} + \tilde{\bB} x_t^2 \bW^\top) \prod_{j=i+1}^{t-1}(\bar{\bA} + \tilde{\bB} x_j^2 \bW^\top) = \mathbf{S}_{t+1, t+k} (2x_t \tilde{\bB} \bW^\top) \mathbf{S}_{i+1, t-1}$. Combining this into the memory expression, we obtain
        \begin{equation*}
            \begin{aligned}
                \mathcal{M}(t+k, t) &= \mathbf{S}_{t+1, t+k} (2x_t \tilde{\bB} \bW^\top) \mathbf{S}_{1, t-1} \bh_0 + \sum_{i=1}^{t-1} \mathbf{S}_{t+1, t+k} (2x_t \tilde{\bB} \bW^\top) \mathbf{S}_{i+1, t-1} \bar{\bB}_i x_i
                + \mathbf{S}_{t+1, t+k} \bar{\bB}_t
            \end{aligned}
        \end{equation*}
    \end{proof}

\section{Proof of Lemma~\ref{lem:stabilityexpression}}
    \repthm{lem:stabilityexpression}
    \begin{proof}
        Plugging the state transition~\eqref{eq:state_transition_eigen} into the unrolled form~\eqref{eq:unrollform_translation}, we obtained the following hidden state expression:
        \begin{equation*}
            \begin{aligned}
                \bh_{t} &=  \left(\prod_{i=1}^t
                (\bar{\bA} + \tilde{\bB} \bx_i \bx_i^\top \bW^\top)\right) \bh_0 + \sum_{i=1}^t \left( \prod_{j=i+1}^{t} (\bar{\bA} + \tilde{\bB} \bx_j \bx_j^\top \bW^\top) \right)
                \bar{\bB}_i \bx_i \\
                &= \bP \prod_{i=1}^t \begin{pmatrix} \lambda_1 & & &\\ & \lambda_2 & & \\ & & \ddots & \\ & & & \lambda_H + \gamma x_{i}^2 \end{pmatrix} \bP^\top \bh_0 + \sum_{i=1}^t \bP \prod_{j=i+1}^t \begin{pmatrix} \lambda_1 & & &\\ & \lambda_2 & & \\ & & \ddots & \\ & & & \lambda_H + \gamma x_{j}^2 \end{pmatrix} \bP^\top \bar{\bB}_i x_i \\
                &= \bP \begin{pmatrix} \lambda_1^t & & &\\ & \lambda_2^t & & \\ & & \ddots & \\ & & & \prod_{i=1}^t (\lambda_H + \gamma x_{i}^2) \end{pmatrix} \bP^\top \bh_0 + \sum_{i=1}^t \bP \begin{pmatrix} \lambda_1^{t-i} & & &\\ & \lambda_2^{t-i} & & \\ & & \ddots & \\ & & & \prod_{j=i+1}^t(\lambda_H + \gamma x_{j}^2) \end{pmatrix} \bP^\top \bar{\bB}_i x_i
            \end{aligned}
        \end{equation*}
        Since $0 \leq |\lambda_1|, \dots, |\lambda_{H-1}| < 1$, therefore, $0 \leq |\lambda_1^t|, \dots, |\lambda_{H-1}^t| < 1$, the terms related to the first $H-1$ eigenvalues are stable regardless of $t$. $\bh_t$ is finite if and only if the terms related to $\lambda_H$ are finite. In particular, the related to $\lambda_H$ in the second part on the right hand side can be rewritten as
        \begin{equation*}
            \sum_{i=1}^t \prod_{j=i+1}^t(\lambda_H + \gamma x_{j}^2) \bP_H \bP_H^\top \bar{\bB}_i x_i.
        \end{equation*}
        For any bounded $\bP_H$ and probabilistically bounded $\bar{\bB}_i x_i$, this is finite if $\sum_{i=1}^t \prod_{j=i+1}^t |\lambda_H + \gamma x_{j}^2|$ is finite, which further indicates $\prod_{i=0}^t |\lambda_H + \gamma x_{i}^2|$ is finite in the first part on the right hand side.
    \end{proof}

\section{Proof of Theorem~\ref{thm:probabilistic_bound_statibility}}
    \begin{lemma}\label{lem:subexponential_log_chisquare}
        Given a standard Gaussian random variable $X$ and real positive scalar $c > 0$, $\log (c + \gamma X^2)$ is a sub-exponential random variable for some positive pairs of $(\nu, b)$:
        \begin{equation}
            \mathbb{E} (e^{\lambda \log(c + \gamma X^2)}) \leq e^{\frac{\nu^2 \lambda^2}{2}} \text{ for } |\lambda| < \frac{1}{b}.
        \end{equation}
    \end{lemma}
    \begin{proof}
        \cite{vershynin2018high} stated that to prove a random variable $Y$ is sub-exponential, it suffices to prove the tail bounds of both sides are sufficiently small, that is,
        \begin{equation}
            P(|Y - \mathbb{E}(Y)| \geq t) \leq e^{-\frac{t}{K_1}}
        \end{equation}
        for all $t \geq 0$ and for some positive value $K_1$. Therefore, we will prove the following two inequalities: 
        \begin{gather}
            P(\log (c + \gamma X^2) - \mathbb{E}(\log (c + \gamma X^2)) \geq t) \leq e^{-\frac{t}{K_1}} \label{eq:log_chisquare_ineq1} \\
            P(-\log (c + \gamma X^2) + \mathbb{E}(\log (c + \gamma X^2)) \geq t) \leq e^{-\frac{t}{K_1}} \label{eq:log_chisquare_ineq2}.
        \end{gather}

        Denote $\mathbb{E}(\log (c + \gamma X^2)) = \mu$, for \eqref{eq:log_chisquare_ineq1},
        \begin{equation*}
            P\bigg(\log (c + \gamma X^2) - \mu \geq t \bigg) = P\bigg(c + \gamma X^2 \geq e^{\mu + t}\bigg) = P\bigg(X^2 \geq \frac{1}{\gamma}(e^{\mu + t} - c)\bigg).
        \end{equation*}
        Since $X^2$ is a chi-square random variable and is sub-exponential, it holds that $P( X^2 \geq t) \leq e^{-\frac{t^2}{K_1}}$. Thus, 
        \begin{equation*}
            P\bigg(\log (c + \gamma X^2) - \mu \geq t \bigg) = P\bigg(X^2 \geq \frac{1}{\gamma}(e^{\mu + t} - c)\bigg) \leq e^{-\frac{(e^{\mu + t} - c)}{ \gamma K_1}}.
        \end{equation*}
        To compare $e^{\mu + t} - c$ and some scaling of $t$, we construct function $f(t) = e^{\mu + t} - c - e^{\mu}t$. The derivative of $f$ satisfies $f'(t) = e^{t+\mu} - e^{\mu} \geq 0$ when $t \geq 0$. And $\mu \geq \log c$ as $\log (c + \gamma X^2) \geq \log c$ for any $X^2$ values, therefore, $f(0) = e^{\mu} - c \geq e^{\log c} - c = 0$. Combining $f(0) \geq 0$ and $f'(t) > 0$, we have $f(t) \geq 0$ for $t \geq 0$. Therefore, $e^{\mu + t} - c \geq e^{\mu} t$ and $e^{-\frac{(e^{\mu + t} - c)}{ \gamma K_1}} \leq e^{-\frac{ t \exp(\mu)}{\gamma K_1}}$. Therefore,
        \begin{equation*}
            P\bigg(\log (c + \gamma X^2) - \mu \geq t \bigg) \leq e^{-\frac{ t \exp(\mu)}{\gamma K_1}}.
        \end{equation*}

        For \eqref{eq:log_chisquare_ineq2},
        \begin{equation*}
            P(-\log (c + \gamma X^2) + \mu \geq t) = P(\log (c + \gamma X^2) \leq \mu - t).
        \end{equation*}
        Because $\log (c + \gamma X^2) \geq \log c$ regardless of the value of chi-square random variable $X^2$, if $t$ is larger than $\mu - \log c$, $P(\log (c + \gamma X^2) \leq \mu - t) \leq P(\log (c + \gamma X^2) \leq \log c) = 0$, the inequality is automatically satisfied.

        When $0 < t < \mu - \log c$, we further derive $P(\log (c + \gamma X^2) \leq \mu - t)$ as
        \begin{equation*}
            P(\log (c + \gamma X^2) \leq \mu - t) = P( X^2 \leq \frac{1}{\gamma}(e^{\mu - t } - c)) = \mathrm{CDF}_{\chi^2}(\frac{1}{\gamma}(e^{\mu - t } - c)),
        \end{equation*}
        where $\mathrm{CDF}_{\chi^2}$ denotes the cumulative distribution function for chi-square random variable with degrees of freedom 1. It can be shown that $\mathrm{CDF}_{\chi^2}(v) < \sqrt{v}$ for any $v \geq 0$. Applying this inequality, 
        \begin{equation*}
            P(\log (c + \gamma X^2) \leq \mu - t) \leq \sqrt{\frac{e^{\mu - t } - c}{\gamma}} = \sqrt{e^{-t} \frac{e^\mu - ce^t}{\gamma}}.
        \end{equation*}
        Using Jensen inequality, $e^{\mu} = e^{\mathbb{E}(\log (c + \gamma X^2))} \leq \mathbb{E}e^{\log (c + \gamma X^2)} = c + \gamma$, which further allows us to rewrite the bound as
        \begin{equation*}
            P(\log (c + \gamma X^2) \leq \mu - t) \leq \sqrt{e^{-t} \frac{c + \gamma - ce^t}{\gamma}} = e^{-t/2} \sqrt{1 - \frac{c(e^t - 1))}{\gamma}} \leq e^{-t/2} \sqrt{1} = e^{-t/2},
        \end{equation*}
        where the last inequality holds as $t \geq 0$ and thus $c(e^t-1) \geq 0$.

        Now we have proved the tail bound in equation~\eqref{eq:log_chisquare_ineq1} and \eqref{eq:log_chisquare_ineq2}, which implies $\log (c + \gamma X^2)$ is a sub-exponential random variable. By definition of sub-exponential random variable, there exists a positive pair $(\nu, b)$ such that
        \begin{equation*}
            \mathbb{E} (e^{\lambda \log(c + \gamma X^2)}) \leq e^{\frac{\nu^2 \lambda^2}{2}} \text{ for } |\lambda| < \frac{1}{b}.
        \end{equation*}
    \end{proof}

    \repthm{thm:probabilistic_bound_statibility}
    \begin{proof}
        Denote random variable $Y = \log \left( \prod_{i=1}^t ( |\lambda_H| + \gamma x_i^2) \right) = \sum_{i=1}^t \log( |\lambda_H| + \gamma x_i^2)$. According to Lemma~\ref{lem:subexponential_log_chisquare}, $(|\lambda_H| + \gamma x_i^2)$ is a sub-exponential random variable with parameters $(\nu, b)$. The moment generating function of $Y$ satisfies
        \begin{equation}
          \begin{aligned}
            \mathbb{E}(e^{u Y}) &= \mathbb{E}(e^{\sum_{i=1}^t u \log( |\lambda_H| + \gamma x_i^2)}) = \mathbb{E}(\prod_{i=1}^t e^{u \log ( |\lambda_H| + \gamma x_i^2) }) = \prod_{i=1}^t \mathbb{E}(e^{u \log (|\lambda_H| + \gamma x_i^2)}) \leq \prod_{i=1}^t e^{\frac{\nu^2 \lambda^2}{2}} \\
            &= e^{\frac{(\sqrt{t}\nu)^2 \lambda^2}{2}}.
          \end{aligned}
        \end{equation}
        Note that $u$ is the parameter of the moment generating function and $\lambda_H$ is the eigenvalue of $\bar{A}$. Hence, $Y$ is also sub-exponential with parameters $(\sqrt{t}\nu, b)$. Applying the sub-exponential tail bound~\cite{wainwright2019_statistics}, we obtain
        \begin{equation}
            P(Y \geq \mathbb{E}Y + z) \leq \left\{ 
            \begin{aligned}
                & e^{-\frac{z^2}{2 t\nu^2}} \quad \text{if } 0 \leq z \leq \frac{t\nu^2}{b} \\
                & e^{-\frac{t}{2b}} \quad \text{if } z > \frac{t\nu^2}{b}.
            \end{aligned} \right.
        \end{equation}

        Now we transform back to $\prod_{i=1}^t (|\lambda_H| + \gamma x_i^2)$ from $Y$ by applying $e^{Y}$. Note that the exponential function is convex, and applying Jensen inequality, we have
        \begin{equation}
            \mathbb{E}(e^Y) \geq e^{\mathbb{E} (Y)}.
        \end{equation}
        Additionally, the expectation $\mathbb{E}(e^Y)$ can be derived as $\mathbb{E}(e^Y) = \mathbb{E}(\prod_{i=1}^t (|\lambda_H| + \gamma x_i^2)) = \prod_{i=1}^t \mathbb{E}(|\lambda_H| + \gamma x_i^2) = (|\lambda_H| + \gamma)^t$. Therefore, we bound $e^Y$ by
        \begin{equation}
            P(\prod_{i=1}^t (|\lambda_H| + \gamma x_i^2) \geq e^z \mathbb{E}(e^Y)) = P(e^Y \geq e^z \mathbb{E}(e^Y))  \leq P(e^Y \geq e^z e^{\mathbb{E}(Y)}) = P(Y \geq \mathbb{E}(Y) + z).
        \end{equation}
        After explicitly writing out $Y$ and $\mathbb{E}(Y)$, we have the following equivalent expression:
        \begin{equation}
            P(\prod_{i=1}^t (|\lambda_H| + \gamma x_i^2) \geq e^z (|\lambda_H| + \gamma)^t) \leq \left\{
            \begin{aligned}
                & e^{-\frac{z^2}{2 t\nu^2}} \quad \text{if } 0 \leq z \leq \frac{t\nu^2}{b} \\
                & e^{-\frac{t}{2b}} \quad \text{if } z > \frac{t\nu^2}{b}.
            \end{aligned} \right.
        \end{equation}

        Finally, we derive the bound of $\prod_{i=1}^t |\lambda_H + \gamma x_{i}^2|$ based on that of $\prod_{i=1}^t (|\lambda_H| + \gamma x_i^2) $. Since $|\lambda_H + \gamma x_{i}^2| \leq | \lambda_H | + \gamma x_i^2$ for any real Gaussian random variable $x_i$, whenever $\prod_{i=1}^t |\lambda_H + \gamma x_i^2| \geq c$ holds for a threshold $c$, $\prod_{i=1}^t (|\lambda_H| + \gamma x_i^2) \geq c$ must hold. Therefore,
        \begin{equation*}
            P(\prod_{i=1}^t |\lambda_H + \gamma x_{i}^2| \geq c) \leq P(\prod_{i=1}^t (|\lambda_H| + \gamma x_i^2) \geq c).
        \end{equation*}
        Consequently,
        \begin{equation}
            P(\prod_{i=1}^t |\lambda_H + \gamma x_{i}^2|  \geq e^z (|\lambda_H| + \gamma)^t) \leq \left\{
            \begin{aligned}
                & e^{-\frac{z^2}{2 t\nu^2}} \quad \text{if } 0 \leq z \leq \frac{t\nu^2}{b} \\
                & e^{-\frac{t}{2b}} \quad \text{if } z > \frac{t\nu^2}{b}.
            \end{aligned} \right.
        \end{equation}
        
    \end{proof}

\section{Derivation of inequality~\eqref{eq:final_bound}}
    Using the formula for sum of a finite geometric series, we rewrite equation~\eqref{eq:final_bound} as
    \begin{equation*}
      \begin{aligned}
        & P\left(\sum_{i=1}^t\prod_{j=i+1}^t |\lambda_H + \gamma x_{j}^2| \leq e^z \frac{1 - (|\lambda_H| + \gamma)^t}{1 - |\lambda_H| - \gamma} \right)  \\
        = &P\left(\sum_{i=1}^{t-1}\prod_{j=i+1}^t |\lambda_H + \gamma x_{j}^2| \leq e^z \sum_{i=1}^{t-1} (|\lambda_H| + \gamma)^i \right) \\
        \geq &\prod_{i=1}^{t-1} P\left(\prod_{j=i+1}^t |\lambda_H + \gamma x_{j}^2| \leq e^z (|\lambda_H| + \gamma)^{t-i} \right)
      \end{aligned}
    \end{equation*}
    The last inequality holds according to FKG inequality under the mild assumption that $x_i$ have non-negative partial correlations. The non-negative partial correlations are common in multiple types of sequential data, for example, nearby pixels have similar colors while distant pixel colors are relatively independent when conditioned on all remaining pixels. To prove the inequality, we simplify the notation as $C_i = e^z (| \lambda_H| + \gamma)^{t-i}$. Denote set $A = \big\{ (x_1^2, \dots, x_t^2): \sum_{i=1}^{t-1} \prod_{j=i+1}^t | \lambda_H = \gamma x_j^2| \leq \sum_{i=1}^{t-1} C_i \big\}$, and set $A_i = \big\{ (x_1^2, \dots, x_t^2): \prod_{j=i+1}^t | \lambda_H = \gamma x_j^2| \leq C_i \big\}$. FKG inequality states that for a log supermodular probability measure $\mu: \mathcal{X} \rightarrow \mathbb{R}$, given two functions $f, g$ defined on space $\mathcal{X}$, if the two functions are monotonic with the same increasing/decreasing trend, then $\mathbb{E}_{\mu}(fg) \geq (\mathbb{E}_{\mu} f) (\mathbb{E}_{\mu} g)$. To apply FKG to our case, we define $\mathcal{X}$ as the sapce of joint chi-square random variables $(x_1^2, \dots, x_t^2)$, function $f((x_1^2, \dots, x_t^2)) = \mathbf{1}_{A_1} = \mathbf{1}_{\prod_{j=2}^t | \lambda_H + \gamma x_j^2| \leq C_1}$ and $g((x_1^2, \dots, x_t^2)) = \mathbf{1}_{A_2} = \mathbf{1}_{\prod_{j=3}^t | \lambda_H + \gamma x_j^2| \leq C_2}$, which are both decreasing over $\mathcal{x}$. According to FKG, $P(A_1 \cap A_2) = \mathbb{E}(fg) \geq (\mathbb{E}f)(\mathbb{E}g) = P(A_1) P(A_2)$. Generalizing the equality to more than two sets, we have $P(A) = P(\cap_i A_i) \geq \prod_i P(A_i)$, which is equivalent to the last inequality.
    
    By Theorem~\ref{thm:probabilistic_bound_statibility}, 
    \begin{equation*}
      \begin{aligned}
        P\left(\prod_{j=i+1}^t |\lambda_H + \gamma x_{j}^2| \leq e^z (|\lambda_H| + \gamma)^{t-i} \right) &= 1 - P\left(\prod_{j=i+1}^t |\lambda_H + \gamma x_{j}^2| \geq e^z (|\lambda_H| + \gamma)^{t-i} \right) \\
        \geq \left\{ \begin{aligned}
            &1 - e^{-\frac{z^2}{2 (t-i)\nu^2}} \quad &\text{if } 0 \leq z \leq \frac{(t-i)\nu^2}{b} \\
            &1 - e^{-\frac{t-i}{2b}} \quad &\text{if } z > \frac{(t-i)\nu^2}{b}.
        \end{aligned} \right.
      \end{aligned}
    \end{equation*}
    Choosing $z$ as a large number that satisfies $z > \frac{t\nu^2}{b}$, we have
    \begin{equation*}
      \begin{aligned}
        \prod_{i=1}^{t-1} P\left(\prod_{j=i+1}^t |\lambda_H + \gamma x_{j}^2| \leq e^z (|\lambda_H| + \gamma)^{t-i} \right) &\geq \prod_{i=1}^{t-1} (1 - e^{-\frac{t-i}{2b}}) = \prod_{i=1}^{t-1} (1 - e^{-\frac{i}{2b}}) \geq 1 - \sum_{i=1}^{t-1} e^{-\frac{i}{2b}} \\
        & \geq 1 - \frac{ e^{-\frac{1}{2b}} }{ 1-e^{-\frac{1}{2b}} },
      \end{aligned}
    \end{equation*}
    where $\frac{1}{b}$ is the maximum $u$ value in the moment generating function $\mathbb{E}(e^{u \log(c + \gamma X^2)})$ such that the MGF converges to a finite value. Under our assumption that $X$ is a Gaussian random variable,
    \begin{equation*}
        \mathbb{E}(e^{u \log(c + \gamma X^2)}) = \mathbb{E}( (c + \gamma X^2)^{u} ).
    \end{equation*}
    Since Gaussian random variables have infinite number of moments, the above expectation converged to finite value for any $\lambda > 0$, i.e., $\frac{1}{b} > 0$. 
    \begin{equation*}
        1 - \frac{ e^{-\frac{1}{2b}} }{ 1-e^{-\frac{1}{2b}} } \rightarrow 1 \; (\text{as } \frac{1}{b} \rightarrow +\infty).
    \end{equation*}


\end{document}